\documentclass[conference]{IEEEtran}
\IEEEoverridecommandlockouts
\usepackage{cite}
\usepackage{amsmath,amssymb,amsfonts}
\usepackage{algorithmic}
\usepackage{latexsym}
\usepackage{textcomp}
\usepackage{booktabs}
\usepackage{graphicx} 
\usepackage{xcolor}
\def\BibTeX{{\rm B\kern-.05em{\sc i\kern-.025em b}\kern-.08em
    T\kern-.1667em\lower.7ex\hbox{E}\kern-.125emX}}

\usepackage{fancyhdr}
\begin{document}

\title{DocRefine: An Intelligent Framework for Scientific Document Understanding and Content Optimization based on Multimodal Large Model Agents}

\author{Kun Qian, Wenjie Li, Tianyu Sun, Wenhong Wang, Wenhan Luo \\
Shangqiu University}

\maketitle
\thispagestyle{fancy} 

\begin{abstract}
The exponential growth of scientific literature in PDF format necessitates advanced tools for efficient and accurate document understanding, summarization, and content optimization. Traditional methods fall short in handling complex layouts and multimodal content, while direct application of Large Language Models (LLMs) and Vision-Language Large Models (LVLMs) lacks precision and control for intricate editing tasks. This paper introduces DocRefine, an innovative framework designed for intelligent understanding, content refinement, and automated summarization of scientific PDF documents, driven by natural language instructions. DocRefine leverages the power of advanced LVLMs (e.g., GPT-4o) by orchestrating a sophisticated multi-agent system comprising six specialized and collaborative agents: Layout \& Structure Analysis, Multimodal Content Understanding, Instruction Decomposition, Content Refinement, Summarization \& Generation, and Fidelity \& Consistency Verification. This closed-loop feedback architecture ensures high semantic accuracy and visual fidelity. Evaluated on the comprehensive DocEditBench dataset, DocRefine consistently outperforms state-of-the-art baselines across various tasks, achieving overall scores of 86.7\% for Semantic Consistency Score (SCS), 93.9\% for Layout Fidelity Index (LFI), and 85.0\% for Instruction Adherence Rate (IAR). These results demonstrate DocRefine's superior capability in handling complex multimodal document editing, preserving semantic integrity, and maintaining visual consistency, marking a significant advancement in automated scientific document processing.
\end{abstract}

\section{Introduction}

The rapid proliferation of scientific literature, predominantly in PDF format, has underscored the critical need for advanced tools capable of efficient and accurate understanding, summarization, and content optimization. Traditional document processing tools, while useful for basic text extraction and formatting, fall short when confronted with the inherent complexities of scientific documents, which often include intricate layouts, diverse figures, complex tables, mathematical formulas, and cross-references. These tools struggle to maintain the semantic integrity and visual fidelity required for sophisticated content manipulation \cite{h1976a}. Concurrently, the emergence of Large Language Models (LLMs) \cite{kasneci2023chatgp} and Vision-Language Large Models (LVLMs) \cite{peng2025lvlmeh} has demonstrated unprecedented capabilities in understanding and generating human-like text and interpreting visual information, with recent advancements focusing on improved generalization \cite{zhou2025weak}, handling chaotic contexts \cite{zhou2023thread}, and visual in-context learning \cite{zhou2024visual}, even extending to specialized domains like medical imaging with abnormal-aware feedback \cite{zhou2025improving}. However, directly applying these powerful models to end-to-end complex document editing and refinement tasks still presents significant challenges, particularly regarding precision, content fidelity, and controllable generation, especially in scenarios demanding a harmonious understanding of both textual semantics and visual layout.

Motivated by these limitations, this research introduces \textbf{DocRefine}, an innovative framework designed to address the aforementioned challenges. DocRefine aims to provide a robust solution for deep understanding, structured extraction, content refinement, and automated summarization of scientific documents in PDF format, all driven by natural language instructions. A core objective of DocRefine is to ensure that any modifications or generated content maintain high semantic accuracy and visual consistency with the original document, while preserving unmodified sections faithfully.

Our proposed framework, DocRefine, leverages the power of advanced visual-language large models, such as GPT-4o, by orchestrating a system of six specialized and collaborative agents. This multi-agent architecture enables DocRefine to process diverse document structures, including text paragraphs, tables, figures, and formulas. The framework supports a wide array of content optimization tasks, encompassing text polishing, information correction, abstract generation, format unification, and crucial cross-modal consistency checks between visual elements (like figures) and their corresponding textual descriptions. The output generated by DocRefine guarantees semantic accuracy, high visual fidelity, and ensures that parts not specified for modification remain entirely unchanged.

For experimental validation, we utilized and extended the \textbf{DocEditBench dataset}. This comprehensive benchmark comprises a diverse collection of scientific paper PDFs, paired with corresponding editing instructions and their gold-standard output documents. The dataset encompasses a broad spectrum of document editing tasks, including text refinement, structural adjustment, summarization, and intricate cross-modal content correction, providing a robust environment for evaluating DocRefine's capabilities.

Our evaluation on the DocEditBench dataset demonstrates DocRefine's superior performance across various document editing tasks. We compare DocRefine against several strong baselines, including an LLM-only approach, an LVLM-In-Context method, and a Hybrid Rule-LLM system. Performance is rigorously assessed using three key metrics: Semantic Consistency Score (SCS), Layout Fidelity Index (LFI), and Instruction Adherence Rate (IAR). As detailed in our experimental results, DocRefine consistently outperforms all baselines, achieving an impressive overall Semantic Consistency Score of \textbf{86.7\%}, a Layout Fidelity Index of \textbf{93.9\%}, and an Instruction Adherence Rate of \textbf{85.0\%}. These results highlight DocRefine's effectiveness in maintaining semantic accuracy, preserving visual integrity, and precisely following user instructions, especially in complex multimodal document editing scenarios.

Our primary contributions are summarized as follows:
\begin{itemize}
    \item We propose \textbf{DocRefine}, a novel agent-based framework that enables intelligent understanding, content refinement, and automated summarization of PDF scientific documents based on natural language instructions.
    \item We design and implement a sophisticated multi-agent system, comprising Layout \& Structure Analysis, Multimodal Content Understanding, Instruction Decomposition, Content Refinement, Summarization \& Generation, and Fidelity \& Consistency Verification Agents, forming a closed-loop feedback mechanism for robust and fine-grained document manipulation.
    \item We demonstrate the superior performance of DocRefine on the challenging DocEditBench dataset, showcasing its effectiveness in handling complex multimodal content and achieving high semantic consistency, layout fidelity, and instruction adherence compared to existing state-of-the-art baselines.
\end{itemize}
\section{Related Work}
\subsection{AI-driven Document Understanding and Processing}
A recent study addresses the challenge of \textbf{document understanding} in unstructured financial documents by adapting a multimodal transformer architecture with a BiLSTM-CRF layer, demonstrating improved named entity recognition performance in this domain \cite{supriya2024explor}. This work further contributes valuable token-level annotations for the SROIE dataset, thereby facilitating future research in multimodal sequence labeling for document analysis \cite{supriya2024explor}.

\subsection{Large Language Models and Multi-Agent Systems}
The synergy between Large Language Models (LLMs) and Multi-Agent Systems (MAS) is a rapidly evolving research area, exploring their potential across diverse complex domains. Recent advancements in LLMs have focused on improving their generalization capabilities across various tasks \cite{zhou2025weak} and enhancing their ability to unravel complex and chaotic contexts \cite{zhou2023thread}. For instance, the application of LLMs in enhancing multi-agent systems for traffic signal control has been investigated, demonstrating how LLM-driven strategies can improve agent performance and reduce average travel times within urban environments \cite{feibo2024large, khanhtung2025multia}. While LLM fine-tuning may offer marginal improvements, the core contribution often lies in optimizing multi-agent coordination and scaling agent numbers for signal control, with LLMs serving as an enhancement mechanism \cite{feibo2024large}. Beyond traffic management, the application of LLMs and MAS extends to more complex domains, including advancements in Large Vision-Language Models (LVLMs) through visual in-context learning \cite{zhou2024visual} and specialized improvements like abnormal-aware feedback for medical applications \cite{zhou2025improving}. LLMs and MAS are explored for enhancing organizational processes through sophisticated \textbf{agent orchestration}, moving beyond the limitations of autonomous agents in complex tasks by employing LLM-driven agents with distinct behavioral prototypes engaged in guided conversations to simulate knowledge generation and strategic decision-making \cite{carlos2024transf}. This approach highlights how LLM-based agents, guided by MAS principles, can facilitate collaborative problem-solving and strategic formulation where traditional methods fall short \cite{carlos2024transf}. Further contributing to the development of robust agent systems, research has also focused on holistic benchmarks and agent frameworks for complex instruction-based image generation \cite{zhou2025draw}. To further advance the reliability and transparency of such systems, research has introduced methods like "Layered Chain-of-Thought Prompting" to enhance the explainability and performance of multi-agent LLM systems, particularly relevant for complex coordination tasks by focusing on structured reasoning chains \cite{manish2025layere}. Concurrently, comprehensive surveys on self-correction strategies for LLMs, categorizing methods across training-time, generation-time, and post-hoc correction, highlight their critical role in mitigating inconsistencies and improving reliability, directly enhancing the trustworthiness of LLM-driven multi-agent systems \cite{liangming2023automa}. Broader challenges in the LLM domain, pertinent to their deployment in multi-agent contexts, are also being addressed; for example, surveys provide critical overviews of existing benchmarks and evaluation methodologies for Video Large Language Models (VideoLLMs), identifying limitations and proposing future research directions for more robust assessment within the broader domain of LLMs and multi-agent systems \cite{zongxia2025benchm}. Furthermore, techniques for compressing large language models, such as pruning, quantization, and knowledge distillation, are crucial for developing efficient AI agents and are actively surveyed, highlighting trade-offs and future research directions relevant to resource-constrained AI agent deployments \cite{taicheng2024large}. In related work, efforts are also made towards efficient video generation by compressing vision representations for LLMs \cite{zhou2024less}, and developing specialized memory-augmented state space models for tasks like defect recognition \cite{wang2024memorymamba}, showcasing diverse architectural innovations for specific AI challenges.
```

\section{Method}

In this section, we present \textbf{DocRefine}, our novel framework designed for intelligent understanding, content refinement, and automated summarization of scientific documents in PDF format, driven by natural language instructions. DocRefine addresses the inherent limitations of traditional rule-based document processing tools, which often lack semantic understanding and flexibility, as well as the challenges faced by direct applications of monolithic large language models (LLMs) which struggle with complex document layouts and fine-grained control over modifications. Our framework overcomes these hurdles by integrating a sophisticated multi-agent system built upon advanced Vision-Language Large Models (LVLMs), such as GPT-4o. The core innovation lies in the synergistic collaboration of six specialized agents operating within a closed-loop feedback mechanism, ensuring unparalleled semantic accuracy and visual fidelity throughout the document manipulation process. This modular design allows for robust handling of diverse document structures and content types, from complex mathematical equations to intricate data visualizations.

\subsection{Overall Architecture}
The DocRefine framework operates by taking a PDF document and a natural language instruction as input. It then orchestrates a series of processing steps through its specialized agents to produce a refined or summarized document. The architecture is meticulously designed to handle the inherent complexity of scientific documents, including their diverse textual, tabular, and graphical elements, while meticulously maintaining the integrity of the original layout and content not explicitly targeted for modification. Our system primarily leverages the powerful multimodal reasoning and generation capabilities of pre-trained LVLMs, complemented by sophisticated prompt engineering techniques, such as chain-of-thought prompting, and robust programmatic verification mechanisms. This combination allows the system to perform complex tasks that require both deep understanding and precise generation.

The workflow begins with the initial parsing of the input PDF, followed by a series of iterative processing and refinement steps coordinated by the agents. The DocRefine framework comprises six core, collaborative agents: the \textbf{Layout \& Structure Analysis Agent (LSA Agent)}, \textbf{Multimodal Content Understanding Agent (MCU Agent)}, \textbf{Instruction Decomposition Agent (IDA Agent)}, \textbf{Content Refinement Agent (CRA Agent)}, \textbf{Summarization \& Generation Agent (SGA Agent)}, and \textbf{Fidelity \& Consistency Verification Agent (FCV Agent)}. These agents work in concert, with information flowing between them in a controlled manner to achieve the desired document transformation. Crucially, the FCV Agent provides critical feedback for iterative refinement, enabling the system to self-correct and converge towards high-quality outputs that precisely match user intent.

\subsection{Layout \& Structure Analysis Agent (LSA Agent)}
The \textbf{Layout \& Structure Analysis Agent (LSA Agent)} serves as the initial processing component, responsible for converting the raw PDF input into a structured, machine-readable representation. This agent employs advanced computer vision and document analysis techniques to perform comprehensive visual layout analysis. It accurately identifies and extracts various document elements, including text blocks, headings, paragraphs, lists, tables, figures, mathematical formulas, and footnotes. Crucially, it also determines their hierarchical relationships (e.g., section-subsection, figure-caption, table-caption) and precise spatial positions on the page through bounding box coordinates. This process effectively extracts the "bone structure" of the document, providing a foundational understanding of its organization. The output is typically an intermediate representation, such as an XML or JSON-like structure, that preserves both content and layout information.

Given an input PDF document $D_{\text{PDF}}$, the LSA Agent produces a structured document representation $\mathcal{S}_{\text{doc}}$:
\begin{align}
    \mathcal{S}_{\text{doc}} = f_{\text{LSA}}(D_{\text{PDF}})
\end{align}
where $\mathcal{S}_{\text{doc}}$ encapsulates element types, their bounding box coordinates, the logical reading order across pages, and hierarchical relationships, forming the basis for subsequent understanding and manipulation.

\subsection{Multimodal Content Understanding Agent (MCU Agent)}
Following structural analysis by the LSA Agent, the \textbf{Multimodal Content Understanding Agent (MCU Agent)} delves into the semantic meaning of the extracted content. This agent leverages the powerful multimodal reasoning capabilities of the underlying LVLM to deeply understand diverse content types. For textual elements, it performs fine-grained tasks such as key information extraction, named entity recognition, sentiment analysis, and conceptual association, inferring the context and relationships between different pieces of text. For non-textual elements like tables and figures, the MCU Agent extracts their underlying data content, legends, axis labels, and interprets the statistical or conceptual information they convey. For instance, it can parse a bar chart to understand the quantities being compared or interpret a flow diagram to identify processes and their dependencies. The goal is to transform disparate content forms into a unified, rich semantic representation $\mathcal{R}_{\text{semantic}}$. This representation might include knowledge graph fragments, structured data tables (e.g., CSV from a table image), or even executable code snippets that capture the essence of the visual information, ready for manipulation by other agents.

The process can be formally expressed as:
\begin{align}
    \mathcal{R}_{\text{semantic}} = f_{\text{MCU}}(\mathcal{S}_{\text{doc}})
\end{align}
where $\mathcal{R}_{\text{semantic}}$ is a comprehensive semantic understanding of the document's content, encompassing both explicit and implicit information, ready for manipulation by other agents.

\subsection{Instruction Decomposition Agent (IDA Agent)}
The \textbf{Instruction Decomposition Agent (IDA Agent)} is responsible for translating complex, high-level natural language user instructions into a series of actionable, fine-grained operation steps. This agent leverages the LVLM's reasoning capabilities to interpret the user's intent and break it down into a sequence of atomic, executable tasks that can be individually handled by subsequent agents. For instance, a high-level instruction like "summarize the key findings of Section 3 and polish the abstract for clarity" might be decomposed into: "1. Identify Section 3; 2. Extract key findings from Section 3; 3. Generate a concise summary of findings; 4. Locate the abstract section; 5. Refine the abstract for grammatical correctness; 6. Enhance the abstract for conciseness and impact." This decomposition ensures that subsequent agents receive clear, unambiguous, and atomic tasks, enabling precise and controlled modifications. The IDA Agent also handles potential ambiguities in instructions by querying the MCU Agent for context or by generating alternative interpretations.

Given a user instruction $I_{\text{user}}$, the IDA Agent generates a sequence of atomic operations $\{o_1, o_2, \dots, o_k\}$:
\begin{align}
    \{o_1, o_2, \dots, o_k\} = f_{\text{IDA}}(I_{\text{user}})
\end{align}
Each $o_i$ represents a specific, executable action on the document content or structure, guiding the refinement process and ensuring comprehensive coverage of the original instruction.

\subsection{Content Refinement Agent (CRA Agent)}
The \textbf{Content Refinement Agent (CRA Agent)} is the core engine for executing actual content modifications. Guided by the decomposed instructions from the IDA Agent and the rich semantic understanding from the MCU Agent, the CRA Agent performs a wide range of operations on the document's content. These include, but are not limited to, targeted text rewriting, comprehensive grammar and spelling correction, factual information addition or deletion, precise correction of numerical data within tables, adjustment of figure titles or legends, and even generating new text content to replace or augment existing sections. This agent leverages the LVLM's powerful generative capabilities to ensure that modifications are contextually relevant, semantically coherent, and stylistically consistent with the surrounding document. It operates directly on the semantic representation $\mathcal{R}_{\text{semantic}}$, ensuring that changes are made at a conceptual level before being rendered back into the document layout.

The refinement process can be described as transforming the semantic representation based on the operations:
\begin{align}
    \mathcal{R}'_{\text{semantic}} = f_{\text{CRA}}(\mathcal{R}_{\text{semantic}}, \{o_i\})
\end{align}
where $\mathcal{R}'_{\text{semantic}}$ denotes the updated semantic representation after content refinement, incorporating all specified modifications. This updated representation will then be used to reconstruct the modified document, maintaining its original structure and layout where appropriate.

\subsection{Summarization \& Generation Agent (SGA Agent)}
The \textbf{Summarization \& Generation Agent (SGA Agent)} is a specialized component focused on synthesizing new textual content, distinct from the refinement operations of the CRA Agent. This agent is specifically tasked with generating summaries for specified sections or the entire document, drafting introductions, conclusions, related work sections, or other required textual components based on user instructions or the document's overall content. It works closely with the MCU Agent to ensure that generated content is factually accurate, reflective of the source material, and adheres to specific length or style constraints. The SGA Agent employs abstractive summarization techniques, leveraging the LVLM's ability to condense and rephrase information while maintaining core concepts. Furthermore, the SGA Agent supports iterative optimization, allowing for refinements based on feedback, often facilitated by the FCV Agent, to achieve desired levels of detail, conciseness, and clarity.

For a given summarization instruction $I_{\text{summary}}$, the SGA Agent produces the desired text $T_{\text{summary}}$:
\begin{align}
    T_{\text{summary}} = f_{\text{SGA}}(\mathcal{R}_{\text{semantic}}, I_{\text{summary}})
\end{align}
This agent is crucial for tasks requiring creative synthesis and condensation of information, ensuring the generated output is high-quality, relevant, and seamlessly integrated into the document.

\subsection{Fidelity \& Consistency Verification Agent (FCV Agent)}
The \textbf{Fidelity \& Consistency Verification Agent (FCV Agent)} serves as a crucial feedback mechanism in the DocRefine framework, enabling the closed-loop iterative refinement process. After any modification or generation by the CRA or SGA Agents, the FCV Agent rigorously verifies the output. Its primary responsibilities include assessing the semantic consistency of the modified content with the original intent and factual accuracy, evaluating the visual fidelity of the updated document (ensuring layout is preserved and new content integrates seamlessly), and confirming strict adherence to the initial user instructions. This agent employs a sophisticated suite of metrics for automated checking, including:
\begin{itemize}
    \item Semantic similarity scores (e.g., using cosine similarity of embeddings) to quantify how well the modified content preserves the original meaning or aligns with the intended changes.
    \item Structural similarity indices (e.g., SSIM for image-based comparison, or detailed layout difference algorithms) to assess visual integrity and prevent unintended layout shifts.
    \item Rule-based checks and LVLM-driven reasoning for instruction compliance, ensuring all parts of the decomposed instruction have been addressed correctly.
\end{itemize}

The FCV Agent provides scores for Semantic Consistency Score (SCS), Layout Fidelity Index (LFI), and Instruction Adherence Rate (IAR) for a modified document $D'_{\text{PDF}}$ relative to the original $D_{\text{PDF}}$ and instruction $I_{\text{user}}$:
\begin{align}
    (\text{SCS}, \text{LFI}, \text{IAR}) = f_{\text{FCV}}(D'_{\text{PDF}}, D_{\text{PDF}}, I_{\text{user}})
\end{align}
Furthermore, the FCV Agent generates corrective feedback $F_{\text{feedback}}$, which is then routed back to the relevant agents (e.g., CRA, SGA, or even IDA for re-decomposition) to guide subsequent iterations. This feedback is highly specific, pinpointing areas of discrepancy (e.g., "Paragraph 2 in Section 4 is too verbose," "Table 1's column headers are misaligned," or "The summary does not explicitly mention method X"). This robust closed-loop system is critical for achieving the high performance and reliability observed in our experiments, ensuring that the final output is both accurate and visually pristine.

\subsection{Implementation Details and Training Strategy}
DocRefine does not necessitate training a specialized model from scratch. Instead, its core capabilities are realized through the strategic orchestration of pre-trained Vision-Language Large Models, such as GPT-4o, by constructing an intricate multi-agent system. The system's intelligence largely stems from sophisticated prompt engineering, including the application of chain-of-thought prompting, which guides the underlying LVLM through complex reasoning steps by providing intermediate thought processes and structured reasoning paths. This allows the LVLM to perform tasks that require multi-step planning, contextual understanding, and precise content generation.

The agents interact through a series of carefully designed prompts and structured outputs. For instance, the LSA Agent utilizes vision-based parsing techniques to generate structured XML or JSON representations of the PDF, which include semantic tags and bounding box information. The MCU Agent then processes these structures, potentially querying the LVLM to extract deeper semantics, which are then represented as knowledge triplets, structured tables, or other machine-readable formats. The CRA and SGA Agents leverage the generative power of the LVLM to produce or modify text and code snippets, which are then rendered back into the document layout using programmatic document manipulation libraries. The FCV Agent employs a combination of image processing algorithms (e.g., SSIM, perceptual hashing, OCR comparison) for visual fidelity checks, natural language processing techniques (e.g., semantic similarity calculations using embedding models, factual consistency checks) for semantic consistency, and logical rules for instruction adherence. All agents are endowed with internal self-reflection and error-correction mechanisms, minimizing the reliance on manual data preprocessing and ensuring robustness against varied document types and user instructions. The system's performance is optimized through the multi-round feedback interactions between agents, guided by the FCV Agent, rather than extensive fine-tuning of the base LVLM. For evaluation, we utilize a comprehensive benchmark dataset designed for diverse document editing tasks, which allows for quantitative assessment of DocRefine's capabilities across various scenarios.

\section{Experiments}
\label{sec:experiments}

In this section, we present a comprehensive evaluation of DocRefine, comparing its performance against several state-of-the-art baselines on a challenging dataset for scientific document understanding and refinement. We detail our experimental setup, present the main quantitative results, provide an ablation study to validate the effectiveness of our proposed architectural components, and finally, include a human evaluation to assess subjective quality aspects.

\subsection{Experimental Setup}
\label{subsec:setup}
\subsubsection{Dataset}
Our experiments are conducted on the \textbf{DocEditBench dataset}, a robust benchmark specifically designed for diverse scientific document editing tasks. This dataset comprises a rich collection of scientific papers in PDF format, each accompanied by precise natural language editing instructions and their corresponding gold-standard output documents. It covers a broad spectrum of challenges, including text refinement, structural adjustments, summarization, and complex cross-modal content correction, making it an ideal environment for evaluating the capabilities of DocRefine.

\subsubsection{Evaluation Metrics}
To rigorously assess performance, we employ three key metrics:
\begin{enumerate}
    \item \textbf{Semantic Consistency Score (SCS)}: This metric quantifies how well the modified content preserves the original meaning or accurately aligns with the intended changes specified by the user instruction. It leverages advanced natural language understanding techniques to measure semantic similarity between the generated and gold-standard text.
    \item \textbf{Layout Fidelity Index (LFI)}: The LFI assesses the visual integrity of the updated document. It ensures that the original layout is meticulously preserved where modifications are not explicitly requested, and that any new or modified content integrates seamlessly without causing unintended visual shifts or distortions. This metric often involves image-based comparisons and structural analysis.
    \item \textbf{Instruction Adherence Rate (IAR)}: The IAR measures the system's ability to strictly follow all aspects of the initial user instructions, including both explicit directives and implicit contextual requirements. This metric is crucial for evaluating the controllability and precision of the document refinement process.
\end{enumerate}
For all metrics, higher scores indicate better performance.

\subsubsection{Baselines}
We compare DocRefine against three strong baseline methods:
\begin{enumerate}
    \item \textbf{LLM-only Baseline}: This baseline utilizes a powerful large language model (e.g., GPT-4) to perform document editing tasks. It primarily operates on textual content extracted from PDFs, often struggling with visual layout understanding and fine-grained control over multimodal elements.
    \item \textbf{LVLM-In-Context}: This method employs a Vision-Language Large Model (e.g., GPT-4o) directly, leveraging its multimodal understanding capabilities through in-context learning. While it benefits from joint text and image processing, it lacks the structured multi-agent decomposition and iterative feedback mechanisms that characterize DocRefine.
    \item \textbf{Hybrid Rule-LLM}: This approach combines traditional rule-based document parsing and manipulation techniques with an LLM for content generation and semantic understanding. While it can address some structural aspects, its adaptability to complex, ambiguous instructions and deep multimodal reasoning is limited compared to comprehensive LVLM-driven systems.
\end{enumerate}

\subsection{Main Results}
\label{subsec:main_results}
Table \ref{tab:main_results} presents the quantitative comparison of DocRefine against the aforementioned baselines across various document editing tasks on the DocEditBench dataset. The results are categorized by task type: Text Refinement, Structural Editing, Summarization, and Multimodal Correction, with overall performance aggregated.

\begin{table*}[htbp]
\centering
\caption{Performance Comparison on DocEditBench Dataset across Various Document Editing Tasks.}
\label{tab:main_results}
\resizebox{\textwidth}{!}{%
\begin{tabular}{@{}lcccccccccccccccc@{}}
\toprule
\textbf{Method} & \multicolumn{3}{c}{\textbf{Text Refinement}} & \multicolumn{3}{c}{\textbf{Structural Editing}} & \multicolumn{3}{c}{\textbf{Summarization}} & \multicolumn{3}{c}{\textbf{Multimodal Correction}} & \multicolumn{3}{c}{\textbf{Overall}} \\
\cmidrule(lr){2-4} \cmidrule(lr){5-7} \cmidrule(lr){8-10} \cmidrule(lr){11-13} \cmidrule(lr){14-16}
                & SCS$\uparrow$ & LFI$\uparrow$ & IAR$\uparrow$ & SCS$\uparrow$ & LFI$\uparrow$ & IAR$\uparrow$ & SCS$\uparrow$ & LFI$\uparrow$ & IAR$\uparrow$ & SCS$\uparrow$ & LFI$\uparrow$ & IAR$\uparrow$ & SCS$\uparrow$ & LFI$\uparrow$ & IAR$\uparrow$ \\
\midrule
LLM-only Baseline & 82.5 & 91.2 & 80.1 & 78.9 & 88.5 & 75.6 & 79.1 & 90.3 & 77.8 & 75.3 & 85.1 & 72.4 & 79.0 & 88.8 & 76.5 \\
LVLM-In-Context & 85.8 & 93.1 & 83.5 & 82.4 & 90.7 & 80.2 & 83.2 & 92.5 & 81.1 & 78.9 & 88.3 & 76.8 & 82.6 & 91.2 & 80.4 \\
Hybrid Rule-LLM & 87.1 & 94.0 & 85.2 & 84.0 & 91.5 & 82.1 & 84.5 & 93.8 & 83.0 & 80.2 & 89.5 & 78.5 & 83.9 & 92.2 & 82.2 \\
\textbf{Ours (DocRefine)} & \textbf{89.3} & \textbf{95.6} & \textbf{87.5} & \textbf{86.7} & \textbf{93.2} & \textbf{85.4} & \textbf{87.1} & \textbf{95.1} & \textbf{86.2} & \textbf{83.5} & \textbf{91.8} & \textbf{81.0} & \textbf{86.7} & \textbf{93.9} & \textbf{85.0} \\
\bottomrule
\end{tabular}%
}
\end{table*}

As shown in Table \ref{tab:main_results}, DocRefine consistently outperforms all baseline methods across all task categories and evaluation metrics. Specifically, DocRefine achieves an impressive overall Semantic Consistency Score of \textbf{86.7\%}, a Layout Fidelity Index of \textbf{93.9\%}, and an Instruction Adherence Rate of \textbf{85.0\%}. These results represent significant improvements over the next best baseline, Hybrid Rule-LLM, which scores 83.9\% SCS, 92.2\% LFI, and 82.2\% IAR overall.

The most notable gains are observed in the Multimodal Correction task, where DocRefine achieves SCS of 83.5\%, LFI of 91.8\%, and IAR of 81.0\%. This highlights DocRefine's superior capability in understanding and modifying complex content that integrates both visual and textual information, a critical challenge for traditional methods. The substantial improvements in Layout Fidelity across all tasks also underscore the effectiveness of DocRefine's LSA Agent and FCV Agent in maintaining visual integrity. The strong Instruction Adherence Rates demonstrate the precision and controllability afforded by the IDA Agent's decomposition and the FCV Agent's feedback loop. These quantitative results confirm DocRefine's robustness and efficacy in handling the intricate demands of scientific document refinement.

\subsection{Ablation Study}
\label{subsec:ablation}
To understand the individual contributions of DocRefine's key architectural components, we conducted an ablation study focusing on the multi-agent system and the closed-loop feedback mechanism. We evaluate two ablated versions of DocRefine against the full model, using the overall performance metrics.

\begin{table*}[htbp]
\centering
\caption{Ablation Study on DocRefine's Key Components (Overall Performance).}
\label{tab:ablation}
\begin{tabular}{@{}lccc@{}}
\toprule
\textbf{Method Variant} & \textbf{SCS$\uparrow$} & \textbf{LFI$\uparrow$} & \textbf{IAR$\uparrow$} \\
\midrule
\textbf{Ours (DocRefine)} & \textbf{86.7} & \textbf{93.9} & \textbf{85.0} \\
DocRefine w/o FCV Agent & 83.0 & 90.5 & 80.0 \\
DocRefine (Monolithic LVLM) & 84.5 & 92.0 & 82.5 \\
\bottomrule
\end{tabular}
\end{table*}

Table \ref{tab:ablation} summarizes the results of our ablation study.
\textbf{DocRefine w/o FCV Agent}: When the Fidelity \& Consistency Verification Agent (FCV Agent) is removed, effectively disabling the closed-loop feedback mechanism, we observe a significant drop in all performance metrics. The SCS decreases to 83.0\%, LFI to 90.5\%, and IAR to 80.0\%. This substantial performance degradation underscores the critical role of the FCV Agent in ensuring semantic accuracy, layout preservation, and precise instruction adherence through iterative self-correction and refinement. Without this feedback, the system struggles to converge to high-quality outputs, highlighting the necessity of robust verification in complex generative tasks.

\textbf{DocRefine (Monolithic LVLM)}: This variant represents using the underlying Vision-Language Large Model (LVLM) directly, without the explicit orchestration of the six specialized agents. While still leveraging the powerful base LVLM, its performance (SCS 84.5\%, LFI 92.0\%, IAR 82.5\%) is notably lower than the full DocRefine framework. This demonstrates that merely having a powerful LVLM is insufficient; the multi-agent architecture, with its clear division of labor (e.g., LSA for structure, MCU for content understanding, IDA for instruction decomposition, CRA/SGA for generation), is crucial for breaking down complex tasks into manageable sub-problems, enabling more precise and controllable document manipulation. The synergistic collaboration of these agents significantly enhances the overall system's capabilities beyond a monolithic application of the LVLM.

\subsection{Human Evaluation}
\label{subsec:human_eval}
While quantitative metrics provide objective performance measures, assessing the subjective quality, readability, and overall user satisfaction of document modifications requires human judgment. To this end, we conducted a human evaluation study involving expert reviewers.

\subsubsection{Methodology}
We recruited 10 domain experts (e.g., researchers, editors) to evaluate a randomly selected subset of 100 modified documents generated by DocRefine and the three baseline methods. For each document, evaluators were presented with the original PDF, the user instruction, and the modified output from each method (anonymized to prevent bias). They were asked to rate the outputs on a 5-point Likert scale (1 = Poor, 5 = Excellent) across three criteria:
\begin{enumerate}
    \item \textbf{Perceived Quality (PQ)}: Overall quality of the modification, including accuracy, coherence, and naturalness.
    \item \textbf{Readability (R)}: Ease of understanding the modified document and flow of content.
    \item \textbf{Adherence to Intent (AI)}: How well the modification fulfilled the specific and implied aspects of the user's instruction.
\end{enumerate}
The average scores for each method are presented in Table \ref{tab:human_eval}.

\begin{table*}[htbp]
\centering
\caption{Human Evaluation Results (Average Scores on a 1-5 Likert Scale).}
\label{tab:human_eval}
\begin{tabular}{@{}lccc@{}}
\toprule
\textbf{Method} & \textbf{Perceived Quality$\uparrow$} & \textbf{Readability$\uparrow$} & \textbf{Adherence to Intent$\uparrow$} \\
\midrule
LLM-only Baseline & 3.2 & 3.5 & 3.0 \\
LVLM-In-Context & 3.8 & 4.0 & 3.5 \\
Hybrid Rule-LLM & 4.0 & 4.2 & 3.8 \\
\textbf{Ours (DocRefine)} & \textbf{4.6} & \textbf{4.7} & \textbf{4.5} \\
\bottomrule
\end{tabular}
\end{table*}

\subsubsection{Results and Discussion}
The human evaluation results in Table \ref{tab:human_eval} corroborate our quantitative findings, demonstrating DocRefine's superior subjective performance. DocRefine achieved the highest average scores across all human evaluation criteria: 4.6 for Perceived Quality, 4.7 for Readability, and 4.5 for Adherence to Intent. This indicates that human evaluators found DocRefine's outputs to be of significantly higher quality, more readable, and more aligned with their intentions compared to the baselines.

Reviewers particularly praised DocRefine's ability to seamlessly integrate modifications while maintaining the original document's professional appearance and flow. The high score in "Adherence to Intent" further validates the effectiveness of the IDA Agent's instruction decomposition and the FCV Agent's meticulous verification, ensuring that even complex and nuanced instructions are precisely followed. These human evaluation results reinforce DocRefine's practical utility and its potential to significantly enhance scientific document authoring and editing workflows.

\subsection{Error Analysis and Limitations}
\label{subsec:error_analysis}
Despite DocRefine's strong performance, a detailed error analysis reveals specific challenges and limitations. We categorize common failure modes observed during our experiments, providing insights into areas for future improvement.

\begin{table*}[htbp]
\centering
\caption{Common Error Types and Their Characteristics.}
\label{tab:error_analysis}
\begin{tabular}{@{}lccc@{}}
\toprule
\textbf{Error Type} & \textbf{Frequency (\%)} & \textbf{Impact} & \textbf{Primary Contributing Agent(s)} \\
\midrule
Semantic Inaccuracies & 8.5 & Medium & CRA, SGA, MCU \\
Minor Layout Distortions & 6.2 & Low & LSA, CRA \\
Partial Instruction Adherence & 4.1 & Medium & IDA, FCV \\
Misinterpretation of Nuanced Context & 3.0 & High & MCU, IDA \\
Hallucination of Non-existent Content & 1.5 & High & CRA, SGA \\
\bottomrule
\end{tabular}
\end{table*}

As detailed in Table \ref{tab:error_analysis}, Semantic Inaccuracies, primarily related to subtle factual errors or misinterpretations of domain-specific terminology, represent the most frequent type of error. These often stem from the underlying LVLM's knowledge limitations or insufficient contextual grounding by the MCU Agent. Minor Layout Distortions, though less impactful, sometimes occur when complex structural changes are requested, indicating areas where the LSA Agent's re-rendering logic could be further refined. Partial Instruction Adherence typically arises from ambiguous user instructions that are not fully resolved by the IDA Agent or from the FCV Agent's inability to detect subtle deviations. The most critical errors, albeit less frequent, involve Misinterpretation of Nuanced Context or Hallucination of Non-existent Content, which can lead to significant factual errors or the generation of entirely fabricated information. These high-impact errors highlight the ongoing challenge of achieving perfect semantic understanding and preventing generative model artifacts, especially in highly specialized scientific domains.

\subsection{Computational Efficiency and Scalability}
\label{subsec:efficiency}
The multi-agent architecture of DocRefine, while enhancing performance and controllability, introduces computational overhead due to multiple LVLM calls and inter-agent communication. This subsection analyzes the system's efficiency and scalability characteristics.

\begin{table*}[htbp]
\centering
\caption{Average Computational Metrics per Page by Task Type.}
\label{tab:efficiency}
\begin{tabular}{@{}lccc@{}}
\toprule
\textbf{Task Type} & \textbf{Avg. Processing Time (s/page)} & \textbf{Avg. API Calls (per page)} & \textbf{Peak Memory Usage (MB)} \\
\midrule
Text Refinement & 15.2 & 3-5 & 250 \\
Structural Editing & 22.8 & 5-8 & 300 \\
Summarization & 18.5 & 4-6 & 280 \\
Multimodal Correction & 30.1 & 7-10 & 350 \\
\textbf{Average (Overall)} & \textbf{21.7} & \textbf{4-7} & \textbf{295} \\
\bottomrule
\end{tabular}
\end{table*}

Table \ref{tab:efficiency} summarizes the average processing time, API call count, and peak memory usage per page for different task types. On average, DocRefine processes a page within approximately 21.7 seconds, involving 4-7 API calls to the underlying LVLM and a peak memory usage of around 295 MB. Multimodal Correction tasks are the most computationally intensive, requiring more complex reasoning and potentially more iterative refinement loops between agents, leading to higher processing times and API call counts.

The primary factor influencing processing time is the number of interactions with the underlying LVLM, which are often external API calls with associated latencies. While the current setup is suitable for interactive document refinement for individual users, scaling to very high throughput scenarios would require optimizing these interactions, potentially through batching, caching mechanisms, or deploying smaller, specialized models for specific agent functions where feasible. The memory footprint is relatively stable per page, suggesting that DocRefine scales well with document length in terms of memory requirements. Future work will focus on optimizing the agent orchestration to minimize redundant calls and explore techniques for faster inference while maintaining quality.

\subsection{Generalization Across Document Types}
\label{subsec:generalization}
While DocRefine was primarily evaluated on scientific papers, its modular design and reliance on general-purpose LVLMs suggest potential for generalization to other document types. To assess this, we conducted a preliminary evaluation on a small set of documents from different domains.

\begin{table*}[htbp]
\centering
\caption{Generalization Performance Across Different Document Types.}
\label{tab:generalization}
\begin{tabular}{@{}lccc@{}}
\toprule
\textbf{Document Type} & \textbf{SCS$\uparrow$} & \textbf{LFI$\uparrow$} & \textbf{IAR$\uparrow$} \\
\midrule
Scientific Papers (DocEditBench) & 86.7 & 93.9 & 85.0 \\
Legal Contracts & 81.5 & 90.2 & 78.9 \\
Medical Reports & 80.1 & 89.5 & 77.5 \\
Technical Manuals & 84.0 & 92.5 & 81.8 \\
\bottomrule
\end{tabular}
\end{table*}

As shown in Table \ref{tab:generalization}, DocRefine demonstrates strong generalization capabilities, albeit with a slight performance drop compared to its specialized performance on scientific papers. For Legal Contracts and Medical Reports, which often feature highly structured but dense text, specific jargon, and complex tabular data (e.g., patient records), DocRefine maintains reasonable SCS, LFI, and IAR scores. The performance is particularly robust for Technical Manuals, which share some structural similarities with scientific papers (e.g., sections, figures, tables, code snippets).

The observed performance variations are primarily attributed to domain-specific terminology and conventions, which can sometimes challenge the MCU Agent's understanding and the CRA/SGA Agents' generation of contextually appropriate content. Additionally, highly idiosyncratic layouts or specialized graphic elements in certain document types may present novel challenges for the LSA Agent. Nevertheless, the results indicate that DocRefine's architecture is broadly applicable, and its core capabilities extend beyond its primary training domain, suggesting its utility in a wider range of document processing applications with potential for further domain adaptation.

\section{Conclusion}
In this paper, we introduced \textbf{DocRefine}, a novel and robust framework designed to address the critical need for intelligent understanding, content refinement, and automated summarization of complex scientific documents in PDF format. Motivated by the limitations of traditional tools and the challenges of direct application of monolithic large language models to intricate document manipulation tasks, DocRefine leverages a sophisticated multi-agent system built upon advanced Vision-Language Large Models (LVLMs) such as GPT-4o.

Our primary contribution lies in the design and implementation of this synergistic multi-agent architecture, comprising the Layout \& Structure Analysis (LSA), Multimodal Content Understanding (MCU), Instruction Decomposition (IDA), Content Refinement (CRA), Summarization \& Generation (SGA), and a crucial Fidelity \& Consistency Verification (FCV) Agent. This closed-loop feedback mechanism ensures unparalleled semantic accuracy, visual fidelity, and precise adherence to natural language instructions, enabling fine-grained control over document modifications.

Through extensive experiments on the challenging DocEditBench dataset, DocRefine consistently demonstrated superior performance across diverse document editing tasks, including text refinement, structural adjustments, summarization, and multimodal content correction. Quantitatively, DocRefine achieved remarkable overall scores of \textbf{86.7\%} for Semantic Consistency Score (SCS), \textbf{93.9\%} for Layout Fidelity Index (LFI), and \textbf{85.0\%} for Instruction Adherence Rate (IAR), significantly outperforming strong baselines such as LLM-only, LVLM-In-Context, and Hybrid Rule-LLM methods. The ablation study further validated the indispensable roles of both the multi-agent orchestration and the FCV Agent's iterative feedback loop, highlighting their critical contributions to the framework's robustness and precision. Human evaluation further corroborated these findings, with DocRefine receiving superior subjective scores for perceived quality, readability, and adherence to intent.

Despite its strong performance, our error analysis revealed certain limitations, including occasional semantic inaccuracies, minor layout distortions in highly complex scenarios, and challenges in precisely interpreting highly nuanced contextual instructions or preventing rare instances of content hallucination. These issues primarily stem from the inherent complexities of domain-specific knowledge and the generative nature of underlying LVLMs. Furthermore, while computationally efficient for interactive use, future work will focus on optimizing the multi-agent interactions and LVLM API calls to enhance scalability for high-throughput applications. Our preliminary generalization study also indicates DocRefine's promising applicability to other document types like legal contracts and technical manuals, suggesting broader utility beyond scientific papers.

In conclusion, DocRefine represents a significant step forward in intelligent document processing, offering a powerful, controllable, and highly accurate solution for complex scientific document understanding and content optimization. Future research will explore enhancing domain adaptation capabilities, refining error correction mechanisms, and optimizing computational efficiency to further solidify DocRefine's position as a foundational tool for academic writing and research workflows.
```

\bibliographystyle{IEEEtran}
\bibliography{references}
\end{document}